\documentclass[letterpaper, 10pt, conference]{ieeeconf}  

\IEEEoverridecommandlockouts                              

\usepackage{graphics} 
\usepackage{epsfig} 
\usepackage{amsmath} 
\usepackage{amssymb}  
\usepackage{ upgreek }
\usepackage[linesnumbered,ruled]{algorithm2e}
\usepackage[utf8]{inputenc}

\usepackage{xcolor}

\usepackage{graphicx}
\usepackage{float}
\usepackage{amsmath}

\usepackage[official]{eurosym}

\usepackage{algpseudocode}
\usepackage{epstopdf}
\usepackage{multicol}
\usepackage[leftcaption]{sidecap}
\usepackage{booktabs}

\usepackage{amssymb}
\usepackage{arydshln}

\usepackage{todonotes}

\usepackage[hidelinks]{hyperref}
\usepackage{paralist}

\usepackage[nolist]{acronym} 

\newacro{vic}[VIC]{Variable Impedance Control}

\newacro{pi2}[PI\textsuperscript{2}]{Policy Improvement with Path Integrals}
\newacro{dmp}[DMP]{Dynamic Movement Primitive}
\newacro{pdmp}[PDMP]{Periodic \ac{dmp}}
\newacro{cmp}[CMP]{Compliant Movement Primitive}
\newacro{seds}[SEDS]{Stable Estimator of Dynamical Systems}
\newacro{sds}[SDS]{Stable Dynamical Systems}
\newacro{ilc}[ILC]{Iterative  Learning  Control}
\newacro{gp}[GP]{Gaussian Process}
\newacro{gpr}[GPR]{Gaussian Process Regression}
\newacro{gmm}[GMM]{Gaussian Mixture Model}
\newacro{gmr}[GMR]{Gaussian Mixture Regression}
\newacro{ppc}[PPC]{Passivity-Preservation Control}
\newacro{tpgmm}[TP-GMM]{Task-Parameterized \ac{gmm}}
\newacro{lfd}[LfD]{Learning from Demonstration}
\newacro{il}[IL]{Imitation Learning}
\newacro{wls}[WLS]{weighted least-squares}
\newacro{spd}[SPD]{Symmetric Positive Definite}
\newacro{emg}[EMG]{Electromyography}
\newacro{vil}[VIL]{Variable Impedance Learning}
\newacro{vilc}[VILC]{Variable Impedance Learning Control}
\newacro{ai}[AI]{Artificial Intelligent}
\newacro{sea}[SEA]{Series Elastic Actuation}
\newacro{dof}[DoF]{Degree of Freedom}
\newacro{vmp}[VMP]{Via-points Movement Primitive}
\newacro{lwr}[LWR]{Locally Weighted Regression}
\newacro{rl}[RL]{Reinforcement Learning}
\newacro{auv}[AUV]{Autonomous Underwater Vehicle}
\newacro{uav}[UAV]{Unmanned Areal Vehicle}
\newacro{cmaes}[CMA-ES]{Covariance Matrix Adaptation-Evolution Strategies}
\newacro{ccdmp}[CC-DMP]{Coordinate Change-\acp{dmp}}
\newacro{gpdmp}[GPDMP]{ Global Parametric Dynamic Movement Primitive}
\newacro{bbo}[BBO]{Black-Box Optimization}
\newacro{dmpbbo}[DMPBBO]{\ac{dmp} \ac{bbo}}
\newacro{ros}[ROS]{Robotic Operating System}
\newacro{cnn}[CNN]{Convolutional Neural Network}
\newacro{nn}[NN]{Neural Network}
\newacro{aedmp}[AEDMP]{AutoEncoded \ac{dmp}}
\newacro{power}[PoWER]{Policy Learning by Weighting Exploration with the Returns}
\newacro{pd}[PD]{Proportional Derivative}
\newacro{momp}[MoMP]{Mixture of Motor Primitives}
\newacro{promp}[ProMP]{Probabilistic Movement Primitives}
\newacro{hrl}[HRL]{Hierarchical \ac{rl}}
\newacro{kmp}[KMP]{Kernelized Movement Primitive}
\newacro{rbf}[RBF]{Radial Basis Function}
\newacro{rbfnn}[RBF-NN]{\ac{rbf}-\ac{nn}}
\newacro{mle}[MLE]{Maximum Likelihood Estimation}
\newacro{qpdmp}[QP-DMP]{Unit Quaternion-based Periodic \ac{dmp}}
\newacro{rmpdmp}[RMP-DMP]{Riemannian Metric-based Periodic \ac{dmp}}
\newacro{}[]{}

\newcommand{\bm}[1]{\boldsymbol{\mathbf{#1}}}

\newcommand{\q}{\bm{q}}

\newcommand{\dq}{\bm{\Dot{q}}}
\newcommand{\LogQ}{\text{Log}^{q}}

\newcommand{\ExpQ}{\text{Exp}^{q}}

\newcommand{\muq}{\bm{\mu}_q}

\newcommand{\trsp}{{^{\top}}}

\newcommand{\figref}[1]{Fig.~\hyperref[#1]{\ref*{#1}}}
\newcommand{\figsref}[1]{Figures~\hyperref[#1]{\ref*{#1}}}
\newcommand{\Figref}[1]{Figure~\hyperref[#1]{\ref*{#1}}}

\newcommand{\tabref}[1]{Table~\hyperref[#1]{\ref*{#1}}}
\newcommand{\secref}[1]{Section~\hyperref[#1]{\ref*{#1}}}
\newcommand{\algoref}[1]{Algorithm~\hyperref[#1]{\ref*{#1}}}

\newcommand{\wrt} {\textit{w.r.t.}~} %
\newcommand{\eg} {\textit{e.g.,}~} %
\newcommand{\ie} {\textit{i.e.,}~} %
\newcommand{\etc}{\textit{etc}} %

\newlength{\Oldarrayrulewidth}

\definecolor{darkgreen}{rgb}{0.0,0.49,0.19}

\newcommand{\panda}{Franka Emika Panda}

\relax
\DeclareFontFamily{U}{eur}{\skewchar\font'177}
\DeclareFontShape{U}{eur}{m}{n}{%
	<-6> eurm5 <6-8> eurm7 <8-> eurm10}{}
\DeclareFontShape{U}{eur}{b}{n}{%
	<-6> eurb5 <6-8> eurb7 <8-> eurb10}{}
\DeclareSymbolFont{ugrf@m}{U}{eur}{m}{n}
\SetSymbolFont{ugrf@m}{bold}{U}{eur}{b}{n}
\DeclareMathSymbol{\upomega}{\mathord}{ugrf@m}{"21}
\graphicspath{{figs/}}

\title{\LARGE \bf
	Periodic DMP formulation for Quaternion Trajectories
}

\author{Fares~J.~Abu-Dakka$^1$,
	Matteo~Saveriano$^2$, and
	Luka~Peternel$^3$
	\thanks{$^1$Intelligent Robotics Group, Dept of Electrical Engineering and Automation, Aalto University, Finland (e-mail:  \texttt{fares.abu-dakka@aalto.fi}).}
	\thanks{$^2$Dept of Computer Science and Digital Science Center, University of Innsbruck, Austria (e-mail:  \texttt{matteo.saveriano@uibk.ac.at}).}
	\thanks{$^3$Delft Haptics Lab, Dept of Cognitive Robotics, Delft University of Technology, Delft, Netherlands (e-mail:  \texttt{l.peternel@tudelft.nl}).}
	\thanks{This work has been partially supported by CHIST-ERA project IPALM (Academy of Finland decision 326304), and by the Austrian Research Foundation (Euregio IPN 86-N30, OLIVER).}
}

\begin{document}
	
	\maketitle
	\thispagestyle{empty}
	\pagestyle{empty}	
	
	\begin{abstract}
		Imitation learning techniques have been used as a way to transfer skills to robots. Among them, dynamic movement primitives (DMPs) have been widely exploited as an effective and an efficient technique to learn and reproduce complex discrete and periodic skills. While DMPs have been properly formulated for learning point-to-point movements for both translation and orientation, periodic ones are missing a formulation to learn the orientation. To address this gap, we propose a novel DMP formulation that enables encoding of periodic orientation trajectories. Within this formulation we develop two approaches: \textit{Riemannian metric-based projection approach} and \textit{unit quaternion based periodic DMP}. Both formulations exploit unit quaternions to represent the orientation. However, the first exploits the properties of Riemannian manifolds to work in the tangent space of the unit sphere. The second encodes directly the unit quaternion trajectory while guaranteeing the unitary norm of the generated quaternions. 
		We validated the technical aspects of the proposed methods in simulation. Then we performed experiments on a real robot to execute daily tasks that involve periodic orientation changes (\ie surface polishing/wiping and liquid mixing by shaking). 
	\end{abstract}
	
	\section{Introduction}
	The ability to control movements and interaction behaviour is one of the fundamental aspects of robots performing their intended tasks. The underlying properties of the task can be encoded and represented by reference trajectories of motion impedance and/or force. Therefore, it is imperative to have a reliable and powerful trajectory encoding method.
	
	One of the most widely used methods is \acfp{dmp} \cite{Ijspeert2002Learning,saveriano2021dynamic}. \acp{dmp} are capable of encoding discrete and periodic trajectories. Discrete trajectories are point-to-point motions with distinct starting and ending point (goal), which are suitable for many daily tasks (\eg pick and place). Nevertheless, many other daily tasks are of periodic nature and periodic trajectories have their own specifics. For example, they have the same starting or ending point and all the neighbouring points have to be smoothly connected to each-other to form a repetitive pattern. To consider these specifics, Ijspeert et al.~\cite{Ijspeert2002Learning} introduced a different formulation for periodic (also called rhythmic) \acp{dmp}.
	Since then, periodic \acp{dmp} and their upgrades have been successfully applied to various realistic tasks of periodic nature, such as surface wiping \cite{gams2016adaptation}, sawing \cite{peternel2018robot}, bolt screwing \cite{peternel2018robotic}, humanoid locomotion \cite{ruckert2013learned} and exoskeleton assistance \cite{peternel2016adaptive}.
	
	\begin{figure}[!t]
		\centering
		\includegraphics[width=0.8\linewidth]{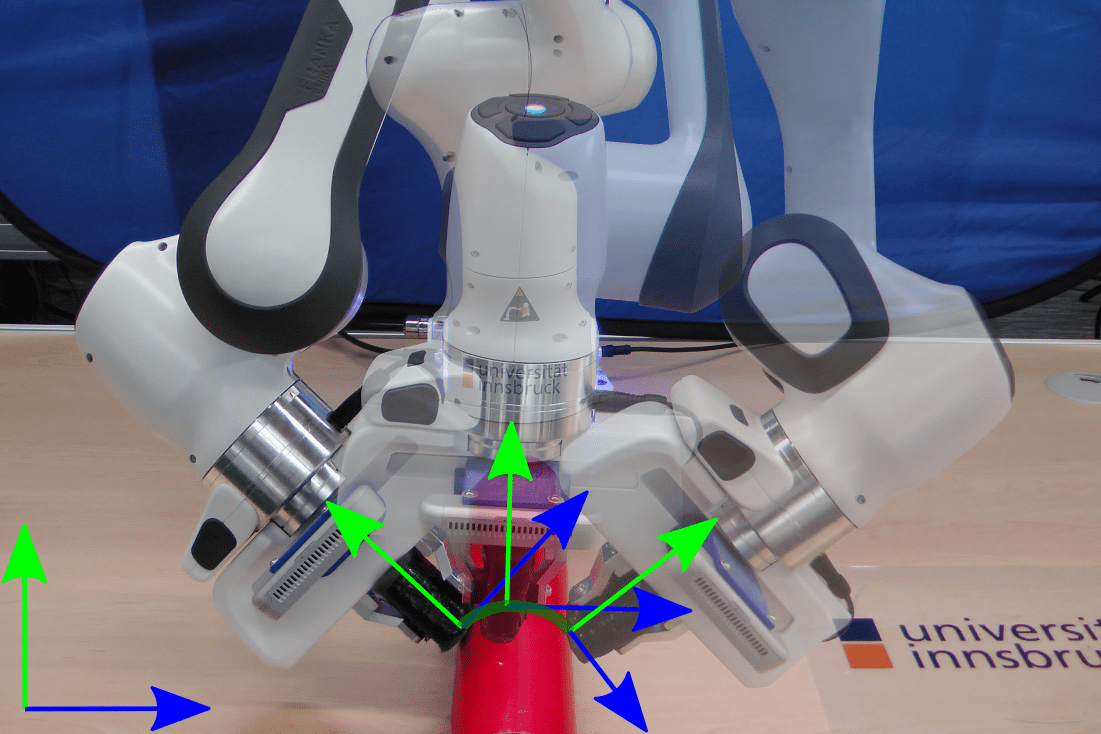}
		\caption{\panda~performs polishing of a curved surface.}
		\label{fig:frontpage}
		\vspace{-4mm}
	\end{figure}
	
	\acp{dmp} were originally designed to encode one \ac{dof} trajectories and are well suited for representing independent signals like joint or Cartesian positions. Synchronization between different \ac{dof} is usually achieved by exploiting a common phase variable. However, in common orientation representations like rotation matrix or unit quaternions the \acp{dof} are interrelated by geometric constraints (\eg unitary norm). To properly consider this constraint both learning and integration need to be carried out by considering the structure of the orientation and perform the integration of rotation elements while obeying its constraint manifold. To account for this interdependence among \acp{dof}, Pastor et al.~\cite{Pastor2011Online} reformulated the original discrete \ac{dmp} to encode unit quaternion profiles. However, this formulation does not take into account the geometry of the unit sphere, since it only used the vector part of the quaternion product. To address this issue \acp{dmp} were reformulated for direct unit quaternion encoding in \cite{AbuDakka2015Adaptation,koutras2020correct}. The quaternion-based \acp{dmp} were also extended to include real-time goal switching mechanism \cite{Ude2014Orientation}. As an alternative to the quaternion representation, a rotation matrix can be used as in~\cite{Ude2014Orientation}. The existing methods are suitable for discrete (point-to-point) orientation trajectories. However, there is no \ac{dmp} approach that can effectively encode periodic orientation trajectories.
	
	
	To address this gap, we propose a novel \ac{dmp} formulation that enables encoding of periodic orientation trajectories. Within this formulation we develop two approaches: \textit{\ac{rmpdmp}} and \textit{\ac{qpdmp}}. 
	The first exploits the fact that the space of unit quaternions is a Riemannian manifold that locally behaves as an Euclidean space (tangent space). Therefore, we can project unit quaternion trajectories onto the tangent space, fit a periodic \ac{dmp} in the tangent space, and project back the output of this \ac{dmp} onto the unit quaternion manifold. The second encodes directly the unit quaternion trajectory and uses quaternion operations to ensure the unitary norm of the integrated quaternions. The proposed approaches are tested on synthetic data and then used to perform surface polishing/wiping and mixing liquids by shaking with a robotic manipulator (see Fig. \ref{fig:frontpage}). 
	
	\section{Related Works}
	\ac{lfd} has been widely used as a convenient way to transfer human skills to robots. It aims at extracting relevant motion patterns from human demonstrations and subsequently applying them to different situations. In the past decades, several \ac{lfd} based approaches have been developed such as: \acp{dmp} \cite{Ijspeert2013Dynamical,AbuDakka2015Adaptation}, \ac{promp} \cite{Paraschos2013}, \ac{sds} \cite{Khansari2011learning, saveriano2020energy}, \ac{gmm} and \ac{tpgmm} \cite{calinon2014}, and \ac{kmp} \cite{Huang2020Toward}. In many previous works, quaternion trajectories are learned and adapted without considering the unit norm constraint (e.g., orientation \ac{dmp} \cite{pastor2009learning} and \ac{tpgmm} \cite{silverio2015learning}), leading to improper quaternions and hence requiring an additional re-normalization.
	
	The problem of learning proper orientations has been investigated in several works. In \cite{Pastor2011Online} was the first attempt to modify the \ac{dmp} formulation to deal with unit quaternions. Their \ac{dmp} considers the vector part of the quaternion product as a measure of the error between the current and the goal orientation. However, this formulation has slow convergence to an attractor point as it is not considering the full geometry constraints of the unit quaternions. Later, the authors in \cite{AbuDakka2015Adaptation, Ude2014Orientation} considered a geometrically consistent orientation error, \ie the angular velocity needed to rotate the current quaternion into the goal one in a unitary time. 
	The stability of both \ac{dmp} formulations is shown in~\cite{saveriano2019merging} using Lyapunov arguments.
	
	In \cite{kim2017gaussian}, \ac{gmm} is used to model the distribution of the quaternion displacements. This probabilistic encoding, together with the exploitation of Riemannian metrics, 
	has been employed in \cite{zeestraten2017approach} to learn orientation trajectories with \ac{tpgmm}.
	
	All previously mentioned approaches exploited \ac{lfd} techniques in order to learn discrete (point-to-point) orientation trajectories and not periodic ones. At the best of the authors knowledge, this is the first work that focuses on learning periodic \ac{dmp} for orientation trajectories. 
	
	The idea of learning a \ac{kmp} in the tangent space of the orientation manifold is presented in~\cite{Huang2020Toward}. While the authors also successfully tested orientation-\ac{kmp} against periodic unit quaternion trajectories, the orientation-\ac{kmp} is an alternative approach and does not resolve the issue within the periodic-\ac{dmp} formulation itself. Due to popularity of \acp{dmp}, there is still a significant functionality gap for many robotic systems that are already based on \acp{dmp} and for future applications for those who would prefer to use \ac{dmp} approach. 
	
	\section{Background}
	\label{sec:background}
	In this section, we provide a brief introduction to the classical periodic \ac{dmp} formulation (sometimes called rhythmic \ac{dmp}). Periodic \acp{dmp} are used when the encoded motion follows a rhythmic pattern. Moreover, we provide basic quaternion notations and operations used in the paper.

	\subsection{Classical Periodic \texorpdfstring{\acs{dmp}}{} formulation}
	The basic idea of periodic \acp{dmp} is to model movements by a system of differential equations that ensure some desired behavior, e.g. convergence to a specified movement cycle in order to encode motion of rhythmic patterns \cite{Ijspeert2002Learning}.
	A \ac{dmp} for a single \ac{dof} periodic trajectory $y$ is defined by the following set of nonlinear differential equations
	\begin{align}
		\dot z &= \Omega \left( {\alpha_z \left( {\beta_z  \left(g - y\right) - z} \right) +f(\phi)} \right),
		\label{eq:dmp_periodic1} \\
		\dot y &= \Omega z,	\label{eq:dmp_periodic2} \\
		\tau\dot{\phi} &= 1, \label{eq:dmp_periodic3}
	\end{align}
	where $g$ is the goal and its value can be set to zero or alternatively to the average of the demonstrated trajectory cycle. $\Omega$ is the frequency and $y$ is the desired periodic trajectory that we want to encode with a \ac{dmp}.
	
	The main difference between periodic \acp{dmp} and point-to-point \acp{dmp} is that the time constant related to trajectory duration is replaced by the frequency of trajectory execution (refer to~\cite{Ijspeert2013Dynamical,Ijspeert2002Learning} for details). In addition, the periodic \acp{dmp} must ensure that the values of the initial phase ($\phi=0$) and the final phase ($\phi=2\pi$) coincide in order to achieve smooth transition during the repetitions. $f(\phi)$ is defined with $N$ Gaussian kernels according to the following equation
	\begin{align}
		f(\phi)&=\frac{{\sum_{i = 1}^N {\Psi_i(\phi) w_i} }}{{\sum_{i = 1}^N {\Psi_i(\phi) } }}\,r, \label{eq:fx_periodic}\\
		\Psi_i(\phi)  &=  \exp\left(h\left( {\cos \left( {\phi  - c_i } \right) - 1} \right)\right),
		\label{eq:psi_periodic}
	\end{align}
	where the weights are uniformly distributed along the phase space, and $r$ is used to modulate the amplitude of the periodic signal~\cite{Ijspeert2002Learning,Gams2009online} (if not used, it can be set to $r=1$~\cite{peternel2016adaptive}).
	
	
	The classical periodic \ac{dmp} described by \eqref{eq:dmp_periodic1}--\eqref{eq:dmp_periodic3} does not encode the transit motion needed to start the periodic one. Transients are important in several applications, \eg in humanoid robot walking, where the first step from a resting position is usually a transient and is needed to start the periodic motion.
	To overcome this limitation, \cite{ernesti2012encoding} modified the classical formulation of periodic \acp{dmp} to explicitly consider transients as motion trajectory that converge towards the limit cycle (\ie periodic) one.

	\subsection{Periodic state estimation and control}
	The phase and frequency of periodic \acp{dmp} can be controlled by an adaptive oscillator, which estimates them as~\cite{Gams2009online}
	\begin{align}
		\dot{\phi} &= \Omega-K\cdot e\cdot\sin(\phi),\label{en:fbl11}\\
		\dot{\Omega} &= -K\cdot e\cdot\sin(\phi),\label{en:fbl21}
	\end{align}
	where $K$ is a positive-value coupling constant and $e = U - \hat{U}$ is a difference between some external signal $U$ (e.g., human muscle activity in exoskeleton control \cite{peternel2016adaptive}) and its internal estimation $\hat{U}$ constructed by a Fourier series~\cite{Petric2011online}
	\begin{align}
		\hat{U} &= \sum_{c=0}^{M}(\alpha_{c}\cos(c\phi)+\beta_{c}\sin(c\phi)), \label{en:fs1}
	\end{align}
	where $M$ is the size of the Fourier series. Fourier series parameters are learnt in the following manner
	\begin{align}
		\dot{\alpha}_{c} &= \eta \cos(c \phi)\cdot e, \label{en:fs2} \\
		\dot{\beta}_{c} &= \eta \sin(c \phi)\cdot e, \label{en:fs3}
	\end{align}
	where parameter $\eta$ is a learning rate. The open parameters can be set to $K=10$, $M=10$ and $\eta=2$ \cite{Petric2011online}. Adaptive oscillators are most useful during online learning to infer periodic state (phase and frequency) in real-time. Nevertheless, they are also useful for offline learning when the recorded signal has variable frequency.

	\subsection{Quaternion operations}
	\label{sec:quat_operation}
	A hyper-complex number is the basis of the quaternion mathematical object.
	Let us define a quaternion as $\q=\nu+\bm{u}\,:\, \nu\in\mathbb{R}, \, \bm{u}=[u_x,u_y,u_z]\trsp \in \mathbb{R}^3,$ $\q \in \mathbb{S}^3$ and $\mathbb{S}^3$ is a unit sphere in $\mathbb{R}^4$. A quaternion with a unit norm is called a unit quaternion and is used to represent an orientation in {3--D} space. In this representation $\q$ and $-\q$ represent the same orientation. $||\q||=\sqrt{\nu^2+u_x^2+u_y^2+u_z^2}$ is the quaternion norm where $||\cdot||$ denotes $\ell_2$ norm. The quaternion conjugation of $\q$ is defined as $\bar{\q}=\nu+(-\bm{u})$, while the multiplication of $\q_1,\q_2\in\mathbb{S}^3$ is defined as 
	\begin{equation}
		\q_1*\q_2 = (\nu_1\nu_2-\bm{u}_1\trsp\bm{u}_2) + (\nu_1\bm{u}_2 + \nu_2\bm{u}_1 + \bm{u}_1 \times \bm{u}_2) 
	\end{equation}
	In order to project unit quaternions back and forth between the unit sphere manifold $\mathbb{S}^3$ and the tangent space $\mathbb{R}^3$ we use logarithmic and exponential mapping operators. $\bm{\zeta} = {\LogQ_{\q_2}(\q_1):\mathbb{S}^3\mapsto\mathbb{R}^3}$ maps $\q_1\in\mathbb{S}^3$ to $\bm{\zeta}\in\mathbb{R}^3$ \wrt to $\q_2$. Consider $\q = \q_1 * \bar{\q}_2$,
	\begin{equation}
		\begin{split}
			\bm{\zeta} =\LogQ_{\q_2}(\q_1) &= \LogQ(\q_1*\bar{\q}_2)=\LogQ(\q)\\
			&=
			\begin{cases}
				\arccos(\nu)\frac{\bm{u}}{||\bm{u||}}, & ||\bm{\zeta||}\neq 0\\
				[0\,\,0\,\,0]\trsp, & \text{otherwise}
			\end{cases}
		\end{split}
		\label{eq:log}
	\end{equation}
	
	Inversely, $\q = {\ExpQ_{\q_2}(\bm{\zeta}) : \mathbb{R}^3\mapsto\mathbb{S}^3}$ maps $\bm{\zeta}\in\mathbb{R}^3$ to $\q\in\mathbb{S}^3$ so that it lies on the geodesic starting point from $\q_2$ in the direction of $\bm{\zeta}$.
	\begin{eqnarray}
		\ExpQ(\bm{\zeta}) &=&
		\begin{cases}
			\cos(||\bm{\zeta||}) + \sin(||\bm{\zeta||})\frac{\bm{\zeta}}{||\bm{\zeta||}}, & ||\bm{\zeta||}\neq 0\\
			1+[0\,\,0\,\,0]\trsp, & \text{otherwise}
		\end{cases}\label{eq:exp}\\
		\q_1 &=& \ExpQ(\bm{\zeta}) * \q_2. \label{eq:expqr}
	\end{eqnarray}
	
	
	In order to ensure discontinuity-free demonstration of an orientation profile and avoid singularity problems, the following two assumptions should be satisfied:
	\paragraph*{Assumption 1} The dot product between each adjacent unit quaternion is greater than zero, such that $\q_t \cdot \q_{t+1}>0$, which guarantees that $\q_t$ and $\q_{t+1}$ are in the same hemisphere. Otherwise, we flip $\q_{t+1}$ such as $\q_{t+1} = \bar{\q}_{t+1}$. 
	\paragraph*{Assumption 2} As discussed in \cite{AbuDakka2015Adaptation,Ude2014Orientation}, the domain of $\LogQ(\cdot)$ extends to all $\mathbb{S}^3$ except $-1+[0\,\,0\,\,0]\trsp$, while the domain of $\ExpQ(\cdot)$ is constrained by $\Vert\bm{\zeta}\Vert<\pi$. Restricting the domain to $\Vert\bm{\zeta}\Vert<\pi$ makes \eqref{eq:log} and \eqref{eq:exp} bijective.

	\section{Proposed Approach}
	\label{sec:proposed}
	Multidimensional periodic variables are encoded using one \ac{dmp} for each \ac{dof} and synchronized by a common phase. This works for variables like joint or Cartesian positions, forces, torques, \etc, where every \ac{dof} of each variable can be encoded and integrated independently form the rest and still reproduce the desired combined behaviour of the robot. However, this is not enough to successfully encode orientations, whose elements are interdependent variables and subject to additional constraints (\ie the orthogonality, the unit norm), without pre- and/or post-processing the data.
	
	In order to overcome this limitation, we propose two approaches to learn periodic orientation movements, namely the \textit{\acf{rmpdmp}} and the \textit{\acf{qpdmp}}.
	In what follows, we consider a periodic demonstration of length $T$ as the trajectory $\bm{\mathcal{Q}}_q=\{\q_t\}_{t=1}^T$, 
	where $\q_t$ are unit quaternions. 
	We assume that the unit quaternions are collected from a single demonstration of a rhythmic pattern and then used to train a periodic \ac{dmp}. 
	
	
	\subsection{\texorpdfstring{\acf{rmpdmp}}{}}
	\label{subsec:riemannian}
	\begin{figure}[!t]
		\centering
		\def\svgwidth{\linewidth}
		{\fontsize{8}{8}
			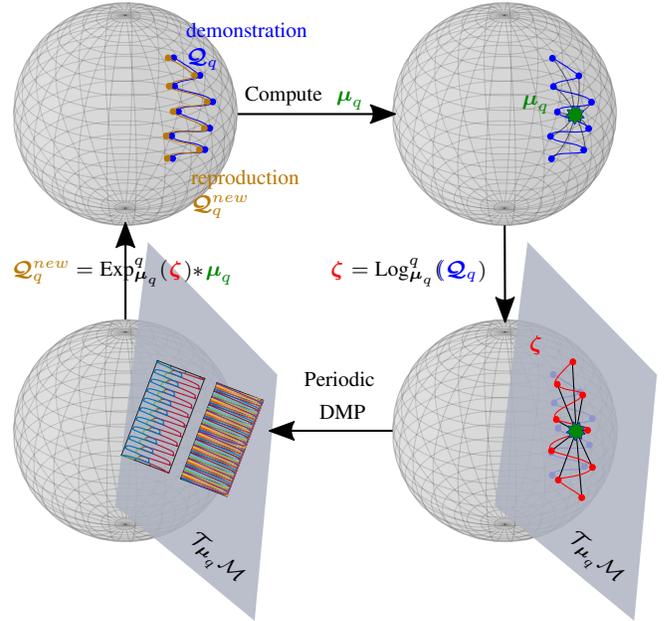}
		\caption{Diagram for the proposed Riemannian metric-based projection for learning and adapting unit quaternion trajectories. The blue data represent unit quaternion demonstration trajectory $\bm{\mathcal{Q}}_q$. The red data represent the projection of $\bm{\mathcal{Q}}_q$ onto the tangent space $\mathcal{T}_{\muq}\mathcal{M}$ \wrt the mean $\muq$. The brown data represent the reproduced unit quaternion trajectory after projecting the reproduced periodic \ac{dmp} trajectory from $\mathcal{T}_{\muq}\mathcal{M}$.}
		\label{fig:framework}
		\vspace{-4mm}
	\end{figure}
	
	A manifold is a topological space that resembles the Euclidean space around each point, \ie it has \textit{locally} the properties of the Euclidean space. A Riemannian manifold $\mathcal{M}$ is a smooth manifold that is equipped with a Riemannian metric. 
	For every point $\bm{m}$ of a manifold $\mathcal{M}$, \ie $\bm{m} \in \mathcal{M}$, it is possible to compute a tangent space $\mathcal{T}_{\bm{m}} \mathcal{M}$, its metric is flat, which allows the use of classical arithmetic tools. In other words, one can perform typical Euclidean operations in the tangent space and project back the result to the manifold. The overall idea is sketched in Fig.~\ref{fig:framework}.
	
	Inspired by~\cite{Huang2020Toward}, we exploit Riemannian operators to project a unit quaternion demonstration $\bm{\mathcal{Q}}_q$ from $\mathcal{M}=\mathbb{S}^3$ onto the tangent space $\mathcal{T}_{\muq}\mathcal{M}$. The tangent space is attached to an auxiliary unit quaternion $\muq$ and each unit quaternion in $\bm{\mathcal{Q}}_q$ is moved to the tangent space by means of $\LogQ_{\muq}(\cdot)$ defined in \eqref{eq:log}.
	Thus, \eqref{eq:log} is used to project $\bm{\mathcal{Q}}_q$ onto $\mathcal{T}_{\muq}\mathcal{M}$ creating $\bm{\mathcal{Q}}_\zeta=\{\bm{\zeta}_t\}_{t=1}^T$. Each point in $\bm{\mathcal{Q}}_\zeta$ belongs to the tangent space and has the same property of $3$D vectors, \ie $\bm{\zeta}_t \in \mathbb{R}^3$. 
	The trajectory $\bm{\mathcal{Q}}_\zeta$ is (numerically) differentiated to compute the 1st and 2nd time-derivatives $\bm{\dot{\zeta}}_t,\bm{\ddot{\zeta}}_t \in \mathbb{R}^3$. The training data $\bm{\zeta}_t,\bm{\dot{\zeta}}_t,\bm{\ddot{\zeta}}_t \in \mathbb{R}^3$ are subsequently encoded using the periodic dynamic system in \eqref{eq:dmp_periodic1}--\eqref{eq:dmp_periodic2}. During the \ac{dmp} execution, at each time-step we use \eqref{eq:exp} and \eqref{eq:expqr} to project back each of the generated $\bm{\zeta}_t^*$ from $\mathcal{T}_{\muq}\mathcal{M}$ onto $\mathbb{S}^3$, creating a new unit quaternion $\q_t^*$.
	
	The center of the tangent space $\muq$ is an open parameter to determine.
	In \cite{Huang2020Toward}, $\muq$ was selected as the initial point of the demonstration. However, this choice cause problems if the extreme orientations in the demonstrated trajectory are close to $0$ or $\pi\,$rad. Using such extreme quaternion as $\muq$ will produce different $\bm{\zeta}$ due to finite precision algebra. This problem may be avoided by selecting $\muq$ as the mean point of demonstrations. This is because the central tendency of the mean---that is the farthest point from the extreme---allows to avoid restricting the input domain, which may be necessary if the initial point is used. In the case of unit quaternions, this mean can be estimated using \ac{mle} \cite{zeestraten2017approach}.
	


	\subsection{\texorpdfstring{\acf{qpdmp}}{}}
	\label{subsec:quatDMP}
	The original formulation of periodic \acp{dmp} were successfully applied to multidimensional independent variables for their individual \ac{dof} $\in \mathbb{R}$. These variables can be joint or Cartesian positions, forces, torques, \etc, where every \ac{dof} of each variable can be encoded and integrated independently form the rest and still reproduce the desired combined behaviour of the robot. However, such formulation is not enough to successfully encode orientations, whose elements are interdependent variables and subject to additional constraints (\ie unitary norm), without pre- and/or post-processing the data.
	
	To properly encode unit quaternions trajectory, inspired by the work on discrete quaternion \acp{dmp}~\cite{AbuDakka2015Adaptation,Ude2014Orientation}, we reformulate the dynamic system in \eqref{eq:dmp_periodic1} and \eqref{eq:dmp_periodic2} as
	\begin{align}
		\bm{\Dot{\eta}} &= \bm{\Omega}\left(\alpha_z(\beta_z2 \, \LogQ(\bm{g}*\overline{\q})-\bm{\eta}) + \bm{f}(\bm{\phi})\right), \label{eq:quat:dw}\\
		\dq &= \bm{\Omega}\frac{1}{2}\bm{\eta}*\q. \label{eq:quat:dq}
	\end{align}
	The unit quaternion $\bm{g} \in \bm{\mathbb{S}}^3$ in~\eqref{eq:quat:dw} denotes the goal orientation and it can be the identity orientation $1+[0\,\,0\,\,0]\trsp$ or the average of the demonstration quaternion profile, which can be estimated using \ac{mle} proposed in \cite{zeestraten2017approach}. $2 \, \LogQ(\bm{g}*\overline{\q})$ defines the angular velocity $\bm{\omega}$ that rotates a unit quaternion $\q$ into $\bm{g}$ within a unit sampling time. $\bm{\Omega}$ is the $3\times 3$ diagonal matrix of frequencies and the nonlinear forcing term is
	\begin{equation}
		\bm{f}(\bm{\phi})=\bm{A}_r\frac{{\sum_{i = 1}^N {\bm{w}_i \Psi_i(\phi)} }}{{\sum_{i = 1}^N {\Psi_i(\phi) } }}, \label{eq:fx_q}
	\end{equation}
	where $\bm{w}_i$ are the weights needed to follow any given rotation profile. We estimate the weights by 
	\begin{equation}
		\begin{split}
		\small
			\frac{{\sum_{i = 1}^N {\bm{w}_i \Psi_i(\phi)} }}{{\sum_{i = 1}^N {\Psi_i(\phi) } }} = \bm{A}_r^{-1} \left(\bm{\Omega}^{-1}\bm{\Dot{\omega}} - (\alpha_z(\beta_z 2 \, \LogQ(\bm{g}*\overline{\q})-\bm{\omega}) )\right),
		\end{split}
	\end{equation}
	where $\bm{A}_r$ is $3\times 3$ diagonal matrix of amplitude modulators.
	The integration of \eqref{eq:quat:dq} is done similarly as in \eqref{eq:expqr}, \ie 
	\begin{equation}
		\q(t+\delta t) = \ExpQ\left(\frac{\delta t}{2}\bm{\Omega}\, \bm{\eta}(t)\right).
	\end{equation}

	\section{Validation}
	\label{sec:validation}
	In order to illustrate the performance of both proposed approaches, we have conducted several examples in simulation as well as in real setup as follows:
	\begin{itemize}
		\item In simulation: (\emph{i}) comparison between both approaches while learning periodic orientation trajectories, and
		(\emph{ii}) coupling the adaptive oscillator to the first approach (\secref{subsec:riemannian}) to control periodic state variables.
		\item Real experiments: (\emph{i}) wiping of uneven surface, and
		(\emph{ii}) mixing of liquids by shaking a bottle.
	\end{itemize}

	\subsection{Simulations}
	\label{sec:simulation}
	
	
	
	\begin{figure}[!t]
		\centering
		\def\svgwidth{\linewidth}
		{\fontsize{8}{8}
			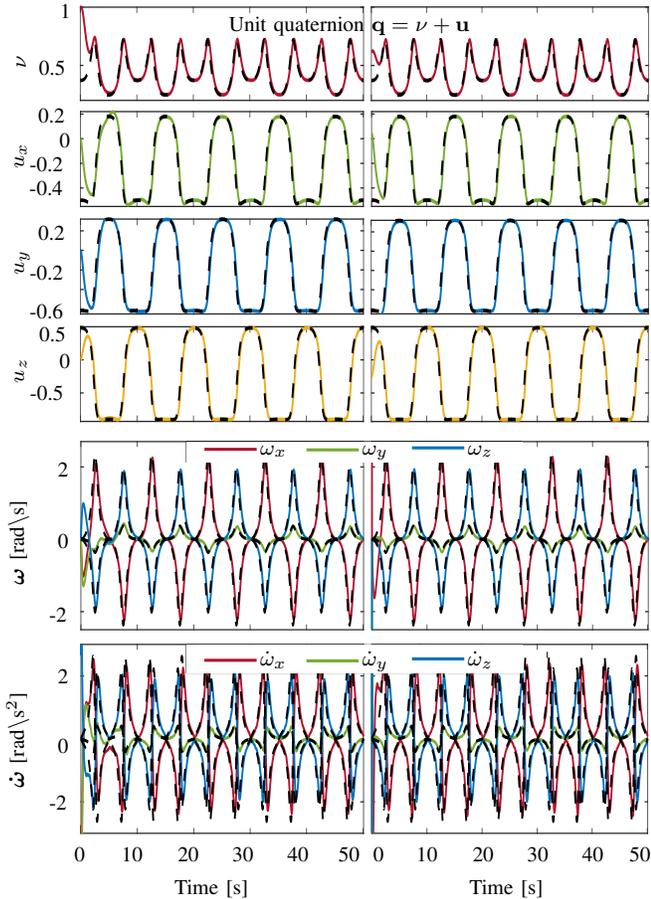}
		\caption{Reproduction of a rhythmic motion using \ac{qpdmp} (left column graphs) and \ac{rmpdmp} (right column graphs). Dashed black lines correspond to the original motion.}
		\label{fig:reproduction:all}
	\end{figure}

	Figure~\ref{fig:reproduction:all} shows the simulation results of using \textit{\ac{qpdmp}} and \textit{\ac{rmpdmp}}. 
	We generated a synthetic periodic orientation trajectory (dashed lines) and use both \ac{dmp} approaches to encode it. Both approaches were able to successfully encode the periodic trajectory with unit quaternions (solid lines). The first four graphs in Fig.~\ref{fig:reproduction:all} show the individual elements of quaternions, while the last two show angular velocity and acceleration, respectively. There is a negligible error between the demonstrated orientation motion and the encoded motion. Also, we did not observe any significant difference between the two \ac{dmp} formulations. In both cases, we initialize the \ac{dmp} state such that the initial orientation differs from the demonstrated one (first four graphs in Fig.~\ref{fig:reproduction:all}). This is to show that the \ac{dmp} effectively converges to a limit cycle that reproduces the demonstration. In this test, we set $\alpha_z=48$, $\lambda = 0.994$, and $\bm{A}_r=\text{diag}\left([1\, 1\, 1]\right)$. 
	
	
	In order to correctly represent the orientation, it is of importance that the norm of the quaternions is unitary. To estimate the ability of the proposed two approaches in keeping this condition, we performed a simulation that tested the norm during the adaptation. 
	
	\begin{figure}[!t]
		\centering
		\def\svgwidth{\linewidth}
		{\fontsize{8}{8}
			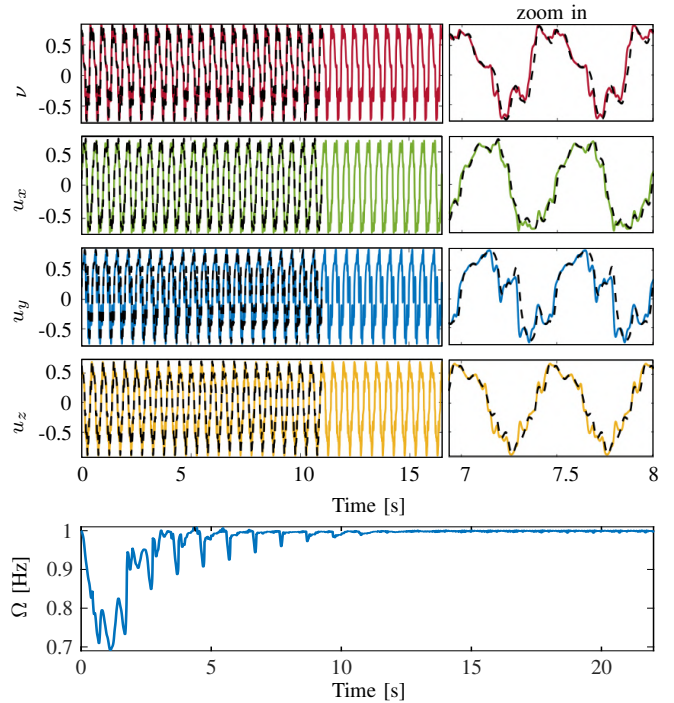}
		\caption{Results of coupling the proposed periodic orientation \acp{dmp} with an adaptive oscillator. The first four graphs show the elements of quaternions as demonstrated (dashed line) and learned (solid line). The frequency of the system is shown in the bottom graph.}
		\label{fig:oscillator}
	\end{figure}
	
	The phase and frequency of the demonstrated periodic motion has to be known in order to be able to learn and encode the \ac{dmp}. However, in practical scenario where the learning is done online, it is difficult to maintain a constant pre-planned frequency in real-time. Therefore, adaptive oscillators can be used to dynamically estimate the phase and frequency of the demonstrated motion in real-time. We performed a simulation to show that the periodic state variables (phase and frequency), as estimated by the adaptive oscillator, can be successfully coupled to the periodic quaternion \acp{dmp} that is operating in the tangent space. The result of the coupling is shown in Fig. \ref{fig:oscillator}. We can see that the system successfully converged to the given periodic frequency, as estimated from the demonstrated input signal.

	\subsection{Experiments}
	\label{sec:experiments}

	We chose surface polishing/wiping and mixing liquids by shaking tasks because they both involve periodic changes of orientation. Many objects are not flat and to polish or wipe them the robot should adapt the orientation periodically in way that the contact with the polishing/wiping tool is perpendicular to the object surface. In case of mixing liquids, we typically change the orientation to shake the container. Therefore, the robot should periodically change the orientation back and forth to stir the liquids inside the container. Both tasks are executed by a $7$ \ac{dof} Franka Emika Panda robot using Cartesian impedance control. In the mixing task, the position is kept fixed at its initial value. 
	
\begin{figure}[!t]
		\centering
		\includegraphics[width=\linewidth]{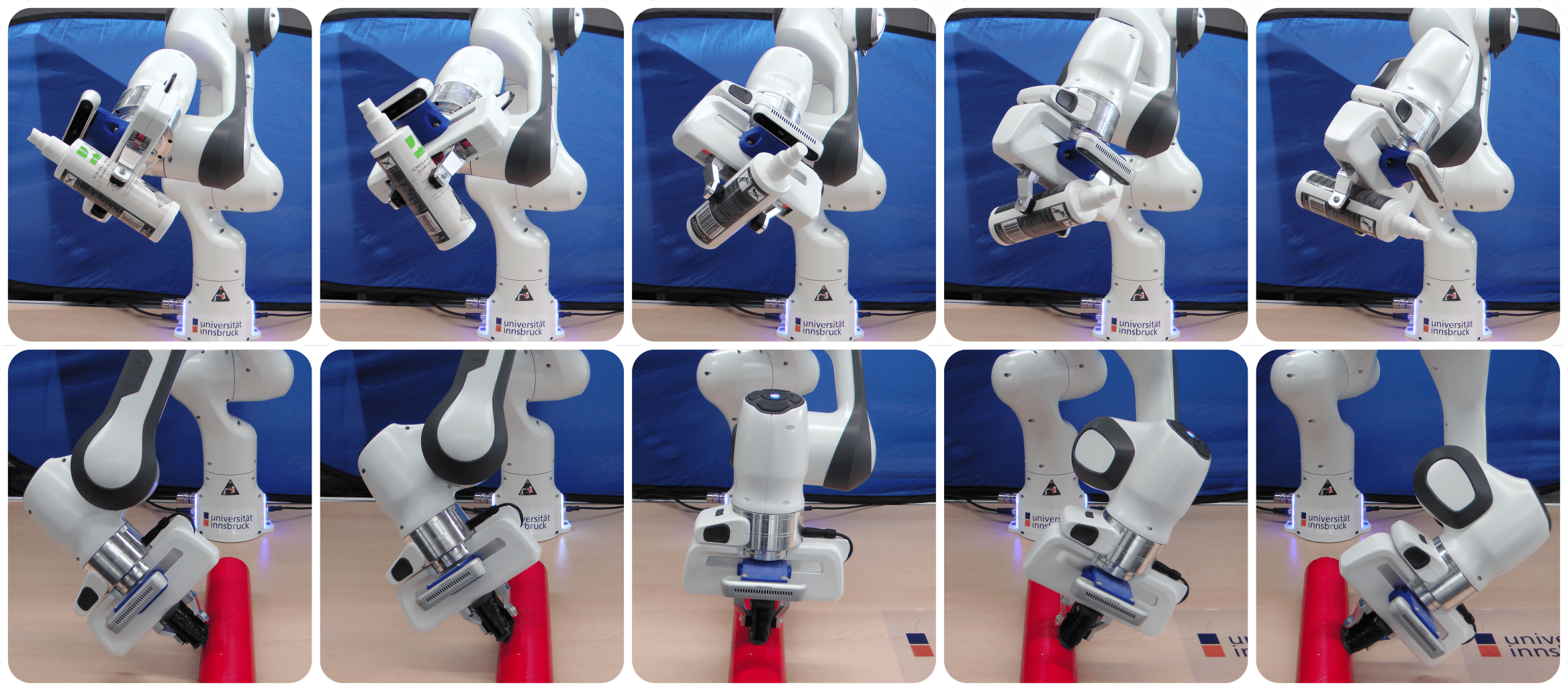}
		\caption{\panda~executes two tasks that require periodic orientation motion with/without periodic position motion: shaking a bottle of liquids (top) and polishing task of curved objects (bottom).}
		\label{fig:polishing}
	\end{figure}

\begin{figure}[!t]
		\centering
		\def\svgwidth{\linewidth}
		{\fontsize{8}{8}
			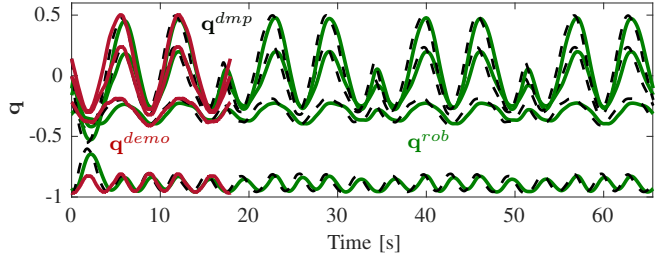}
		\caption{The response of the proposed \ac{dmp} for periodic orientation during the execution of the shaking task. $\q^{rob}$ is actual robot orientation trajectory, $\q^{dmp}$ is the reference orientation trajectory reproduced by our approach, while $\q^{demo}$ is the orientation trajectory used to demonstrate the task.}
		\label{fig:shakingEx}
	\end{figure}

	The series of photos in top-row of Fig. \ref{fig:polishing} shows the robot performing an object shaking task, which is important for mixing various liquids. To perform this task the robot had to periodically change the orientation of the endpoint around a certain point, while holding the container with liquids, in order to produce rapid changes in acceleration (i.e., jerk). Like in the previous tasks, we used the proposed \ac{dmp} method to encode the appropriate orientation modulating trajectories in order to successfully execute the object shaking task. The demonstrated orientation trajectory and the encoded one are shown in Fig. \ref{fig:shakingEx}. We can see that the desired trajectory was reproduced well by the robot during the experiment.

\begin{figure}[!t]
		\centering
		\def\svgwidth{\linewidth}
		{\fontsize{8}{8}
			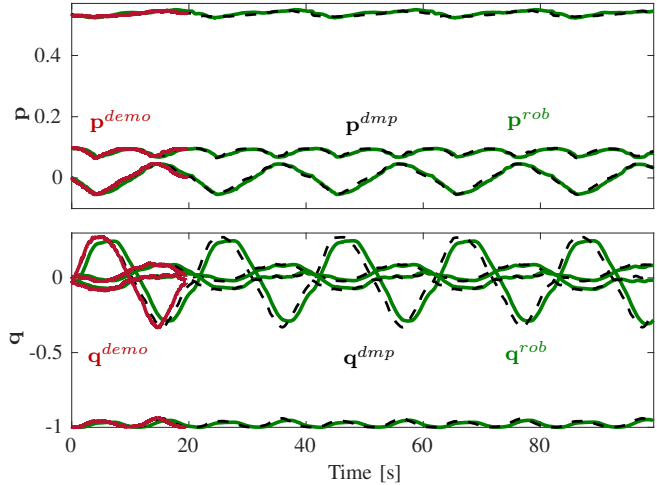}
		\caption{The response of the proposed \ac{dmp} for periodic orientation while executing the polishing/wiping task. $\q^{rob}$, $\q^{dmp}$, and $\q^{demo}$ are defined as in Fig.~\ref{fig:shakingEx}. $\bm{p} = [p_x \, p_y \, p_z]\trsp$ is the position trajectory of the end-effector.}
		\label{fig:polishingEx}
	\end{figure}

	In a different scenario, bottom-row of Fig. \ref{fig:polishing} shows the robot performing polishing/wiping of a highly curved surface. Such task requires the robot to periodically change its end-effector orientation and position in order to keep the contact with the surface of the object. We used the proposed \ac{dmp} method to encode the appropriate orientation modulating trajectories in order to successfully execute the task. The demonstrated and encoded periodic orientation trajectories are shown in Fig. \ref{fig:polishingEx}. 

	\section{Conclusions}
	\label{sec:concl}
	In this paper, we proposed two new approaches to learn periodic orientation motion, represented by unit quaternions, using \ac{dmp}. In the first approach, we exploited Riemannian metrics and operators to perform the learning online on the tangent space of $\mathbb{S}^3$. In the second one, we reformulate the periodic \ac{dmp} equations to directly learn and integrate the unit quaternions online. The performance of both approaches has been validated in simulation as well as in real setups. 
	
	In our experiments, we did not observe a significant outperforming of one of the approaches, as they both behaved similarly. However, the \ac{rmpdmp} has a simpler implementation as it only requires to map the training data onto the tangent space and learn a classical \ac{dmp} there. In other words, one does not have to change the underlying \ac{dmp} formulation and can reuse existing implementations. Moreover, working on the tangent space is a general approach that can be potentially applied to other Riemannian manifolds. Extending the periodic \ac{dmp} formulation to different Riemannian manifolds is the focus of our future research.
	
	\bibliographystyle{IEEEtran}
	\bibliography{ref}

\end{document}